\documentclass{article}

\usepackage{PRIMEarxiv}
\usepackage{helvet}  % DO NOT CHANGE THIS
\usepackage{courier}  % DO NOT CHANGE THIS
\usepackage[hyphens]{url}  % DO NOT CHANGE THIS
\usepackage{graphicx} % DO NOT CHANGE THIS
\usepackage{caption} % DO NOT CHANGE THIS AND DO NOT ADD ANY OPTIONS TO IT
\usepackage{amsfonts}
\usepackage{multirow}
\usepackage{bbding}
\usepackage{threeparttable}
\usepackage[utf8]{inputenc} % allow utf-8 input
\usepackage[T1]{fontenc}    % use 8-bit T1 fonts    
\usepackage{url}            % simple URL typesetting
\usepackage{booktabs}       % professional-quality tables
\usepackage{amsfonts}       % blackboard math symbols
\usepackage{nicefrac}       % compact symbols for 1/2, etc.
\usepackage{microtype}      % microtypography
\usepackage{lipsum}
\usepackage{fancyhdr}       % header
\usepackage{graphicx}       % graphics
\graphicspath{{media/}}     % organize your images and other figures under media/ folder
\usepackage{algorithm}
\usepackage{algorithmic}

\title{SSA: Semantic Structure Aware Inference for Weakly Pixel-Wise Dense Predictions without Cost}

\author{
  Yanpeng Sun \\
  Nanjing University Of Science And Technology \\
  Nanjing, Jiangsu, China\\
  \texttt{yanpeng\_sun@njust.edu.cn} \\
   \And
  Zechao Li* \\
  Nanjing University Of Science And Technology \\
  Nanjing, Jiangsu, China\\
  \texttt{zechao.li@njust.edu.cn} \\
}

\begin{document}
\maketitle

\begin{abstract}
The pixel-wise dense prediction tasks based on weakly supervisions currently use Class Attention Maps (CAM) to generate pseudo masks as ground-truth. However, the existing methods typically depend on the painstaking training modules, which may bring in grinding computational overhead and complex training procedures. In this work, the  semantic structure aware inference (SSA) is proposed to explore the semantic structure information hidden in different stages of the CNN-based network to generate high-quality CAM in the model inference. Specifically, the semantic structure modeling module (SSM) is first proposed to generate the class-agnostic semantic correlation representation, where each item denotes the affinity degree between one category of objects and all the others. Then the structured feature representation is explored to polish an immature CAM via the dot product operation. Finally, the polished CAMs from different backbone stages are fused as the output. The proposed method has the advantage of no parameters and does not need to be trained. Therefore, it can be applied to a wide range of weakly-supervised pixel-wise dense prediction tasks. Experimental results on both weakly-supervised object localization and weakly-supervised semantic segmentation tasks demonstrate the effectiveness of the proposed method, which achieves the new state-of-the-art results on these two tasks.
\end{abstract}

% keywords can be removed

\section{Introduction}
Class Attention Maps (CAM) serves as an essential technology in a wide range of weakly-supervised pixel-wise dense prediction tasks \cite{zhang2020rethinking,zhang2018adversarial}, eg, object location (OL), instance segmentation and semantic segmentation (SS). CAM is proposed to highlight the class-related activation regions for an image classification network, where feature positions related to the specific object class are activated and have higher scores while other regions are suppressed and have lower scores \cite{zhou2016learning}. For specific visual tasks, CAM can be used to infer the object bounding boxes in weakly-supervised OL (WSOL) and generate pseudo-masks of training images in weakly-supervised SS (WSSS). Therefore, obtaining the high-quality CAM is very important to improve the recognition performance of weakly-supervised pixel-wise dense prediction tasks. 
%----------------------------------------------------
\begin{figure}[t]
\centering
\includegraphics[width=0.7\linewidth]{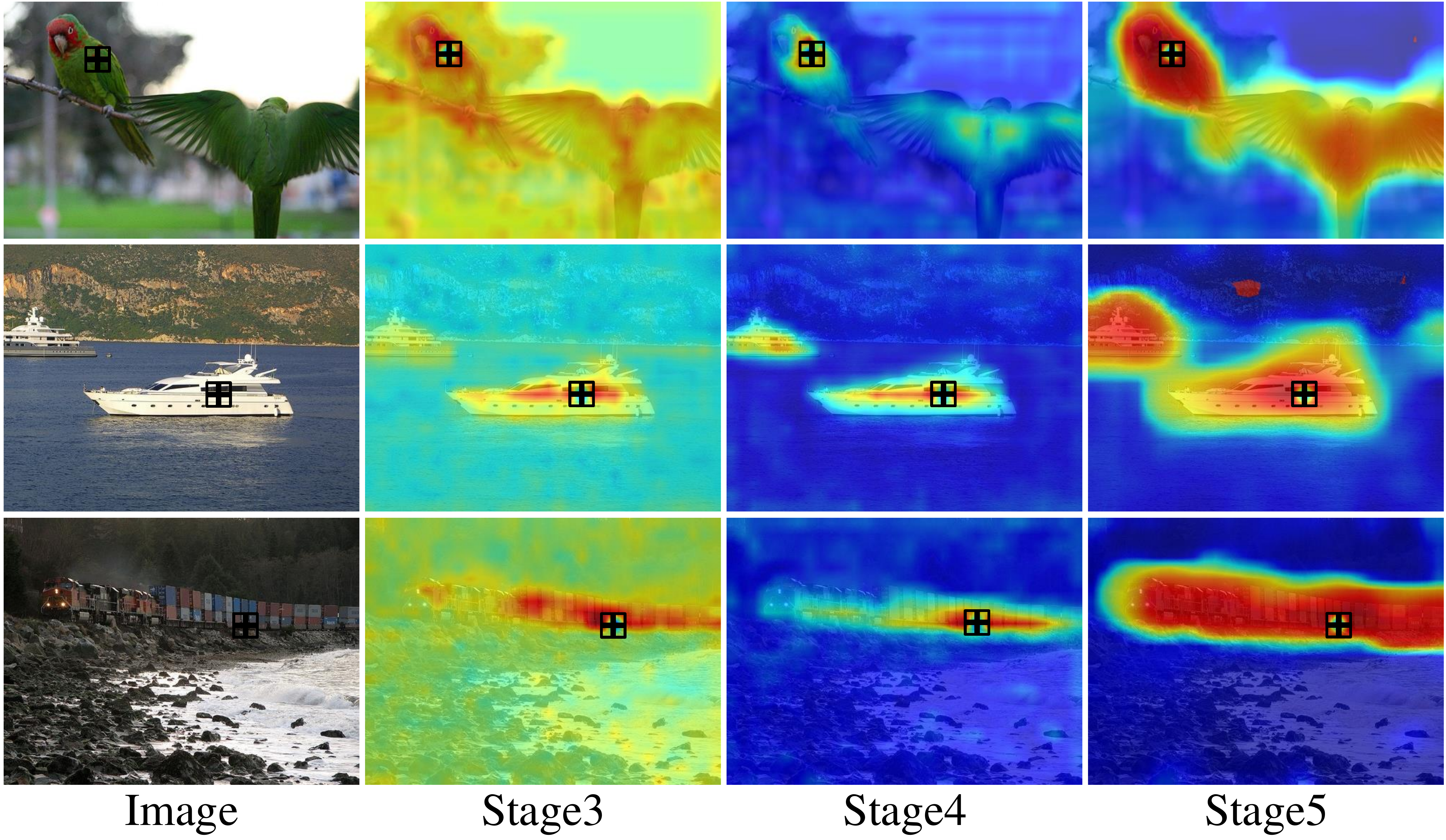} 
\caption{Visualizations of the semantic structure information in backbone stages. The pixels of the same class as the marked pixel are brightly colored. The brighter the color, the higher the similarity. Our motivation comes from this phenomenon.}
\label{point_heatmap}
\vspace{-1.5em}
\end{figure}
%----------------------------------------------------

Recently, considering that the representation ability of the original CAM (ie, generated by a trained convolutional neural network) is immature and not applicable to complex scenarios, gradient-based methods \cite{selvaraju2017grad} and erase-based methods~\cite{wang2020ss} are proposed to improve the CAM quality. Specifically, the gradient-based methods~\cite{chattopadhay2018grad,selvaraju2017grad} obtain CAM by calculating the gradient of the specific image categories relative to deep feature maps. For example, Grad-CAM~\cite{selvaraju2017grad} can extend CAM to various CNN architectures. Although the above methods have achieved satisfied results in some certain scenarios, the gradient-based methods can only capture the most discriminative regions in image, and cannot obtain the detailed features from the image.

The erase-based methods are based on the erasure operation, which improve the localization ability by improving the generalization ability of feature maps~\cite{choe2019attention,zhang2018self}. They remove parts from feature maps or images, and then force the learning model to pay more attention to the remaining second-order regions to improve the CAM integrality.  For example, Cutout~\cite{zhong2020random} randomly erases rectangular area in the image. However, the erase-based methods often introduce irrelevant regions into the training step and may result in the problem of false positives. In addition, the erased termination conditions are not unique under different datasets, which makes it difficult to generalize to multiple scenarios.

Besides, there are also some common key shortcomings. First, the existing methods are implemented in the model training step, which increases not only the training difficulty of the network, but also the parameters of model. The other common shortcoming is about hyperparameters, because different datasets tend to require different hyperparameters. These shortcomings limit the range of applications of CAM in weakly-supervised tasks. Towards this end, this work aims to design a simple yet efficient method to expand CAM. Rethinking the classification network, to improve the probability of identifying objects, pixels belonging to the same category in the feature map have similar representations. To verify this assumption, as shown in Figure~\ref{point_heatmap}, we randomly select one pixel in feature maps generated by different backbone stages to visualize the correlations with other pixels. It can be observed that as the network deepens, the correlation between pixels of the same category in the feature map is stronger. The above visualization results provide strong evidences for our hypothesis and the semantic correlation between pixels is defined as semantic structure information. 

%It should be noted that this assumption is not found in our paper, but existed in previous weakly-supervised pixel-wise dense prediction tasks. This is why the weakly-supervised visual task works.

\begin{figure*}[t]
\centering
\includegraphics[width=\textwidth]{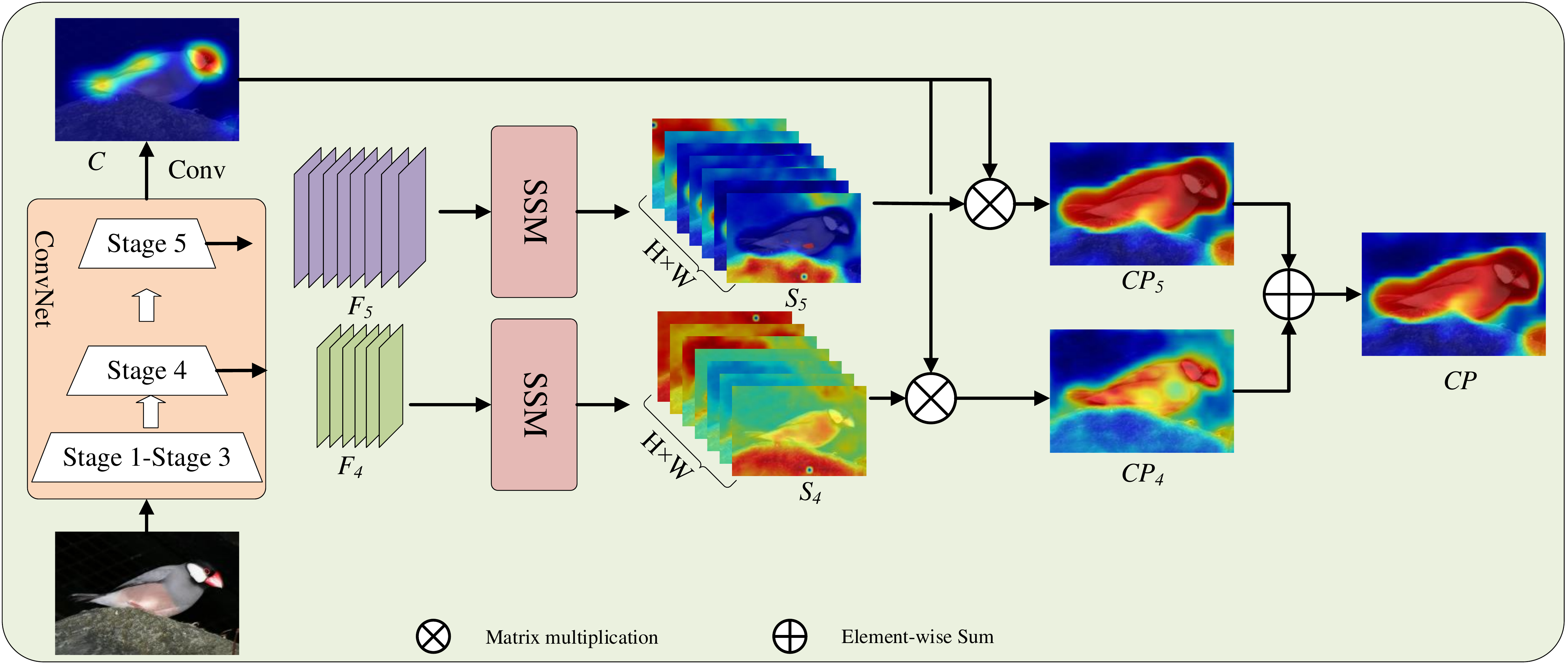} 
\caption{The overall network architecture of the proposed semantic structure aware inference (SSA). Since SSA is only used in the inference CAM stage, it is suitable for all CNN-based models.}
\label{zong}
\end{figure*}

Towards this end, this paper proposes a semantic structure aware inference (SSA) model by leveraging different scales of semantic structure information to generate high-quality CAM, and hence improve the recognition performance of downstream tasks. SSA is introduced in the model inference without any training cost. The overall network architecture is shown in Figure~\ref{zong}. Specifically, a seed CAM is first obtained by using the standard image classification network. Then, the semantic structure modeling module (SSM) is proposed and deployed on different backbone stages to generate the semantic relevance representation. After that, the obtained structured feature representations are used to polish the seed CAM via the dot product operation. Finally, the polished CAMs from different backbone stages are fused as the final CAM. To our best knowledge, this is the first work to improve the quality of CAM without parameters in the model inference step. Experimental results on both WSOL and WSSS demonstrate that SSA can achieve new state-of-the-art performance.

%----------------------------------------------------
%----------------------------------------------------
% The highlighted areas in the second column of images have no semantic relevance to the marked points, while the highlighted areas in the third and fourth columns of images have the same semantics as the marked points. It not only verifies our conjecture, but also proves that there are semantic structures of different scales in deep convolution networks. 

% as shown in Figure \ref{zong}. First, the seed CAM $C$ will be obtained by using the classification network, and then the semantic structure S4 and S5 from $F_4$ and $F_5$ by using the parameter-free semantic structure extractor (SSM). Next, the seed CAM will be updated by using the semantic structure of different stages to get smoothed CAM (CAMS4 and CAMS5). The final CAM is obtained by fusiC. Extensive experiments are conducted on Weakly object localization and weakly semantic segmentation task. The superior performance  Compared with other methods, CAM Smooth produces a more visually clear heatmaps with better object positioning results in a given input image. The superior performance of CAM Smooth is demonstrated in weakly object localization and weakly semantic segmentation.

The main contributions are summarized as follows:
\begin{itemize}
\item This work proposes a novel parameter-free method, termed as semantic structure aware inference (SSA), to improve the CAM quality. It is worth noting that the proposed SSA can be seamlessly integrated into most weakly supervised visual models to improve the performance of the model.
\item The semantic structure modeling module is leveraged to extract the semantic structure information hidden in the feature map, and clearly uncover the semantic correlation between pixels.
\item SSA achieves the new state-of-the-art performance in WSOL and WSSS.
\end{itemize}

\section{Related Work}
\subsection{The Gradient-based Methods}
The gradient-based method \cite{chattopadhay2018grad,fu2020axiom} generates a class activation maps by combining the back-propagation gradient with the feature map. Among them, the representative Grad-CAM~\cite{selvaraju2017grad} uses training weights derived from the global average gradients to obtain CAM. Based on Grad-CAM, Grad-CAM++~\cite{chattopadhay2018grad} improves the accuracy of object localization by introducing the second-order and the third-order gradients. Besides, XGrad-CAM~\cite{fu2020axiom} further introduces two axioms and uses the weighted average gradients to obtain more object details. Smooth Grad-CAM++ \cite{omeiza2019smooth} introduces gradient smoothing into Grad-CAM++, which improves the ability of visual positioning and class object capture. Group-CAM \cite{zhang2021group} designes a noise reduction strategy to filter the less important pixels in the initial masks and replace the unreserved area of the mask with the blurry information. The aforementioned methods explore an efficient way to generate CAM. However the CAM obtained by these methods only identify the most discriminative regions of the object. To improve the accuracy of the CAM, the proposed SSA explores the semantic structure information in the network to expand the CAM.
 
\subsection{The Erase-baesd Methods}

Recently years, some methods \cite{wei2017object, zhong2020random} improve the generalization ability of the classification network by erasing part of the image, forcing the classification network to recognize objects through other areas. Starting with the subject of implementation, erase-based methods can be divided into image-level ones and feature-level ones. Image-level methods first erase part of the image, and then force the network to recognize objects through other non-determinative regions. For example, Random Erasing \cite{zhong2020random} and Cutout \cite{devries2017improved} randomly erase part of the image. ACoL \cite{zhang2018adversarial} erases the most discriminative area in the image. AE-PSL~\cite{wei2017object} erases the area of the CAM from the original image. Feature-level methods dropout units in feature maps during the model training process to prompt the network to focus on other object areas. For example, FickleNet~\cite{lee2019ficklenet} randomly selects hidden feature units during the training phase. ADL \cite{choe2019attention} uses spatial attention map to hide feature units. These methods not only always require additional computing resources in the training phase, but also introduce many hyper-parameters in the inference phase, which limits their application. Therefore, this paper designs a non-parameter SSA strategy in the inference stage to obtain high-quality CAM without cost.

\section{Methodology}
This section first introduces the process of obtaining seed CAM and the re-discussion of CAM, then elaborates the process of obtaining the semantic structure by SSM, and finally elaborates the detailed process of SSA.

\subsection{Rethinking Class Activation Map}
The original CAM is based on a convolutional neural network with global average pooling (GAP). In the inference step, the seed CAM is obtained by weighting the feature map of the last convolution layer output with the weights of the last classification layer. Given an image, the seed CAM of ground-truth class $m$ is computed by:

%In the training step, classification networks use Global Average Pooling (GAP) to integrate the spatial information of the feature map from last CNN layer, and then use the last classification layer to make category prediction. In the inference step, the seed CAM is obtained by weighting the feature map of the last convolution layer output with the weights of the last classification layer. Given an image, the seed CAM of ground-truth class $m$ is computed by:
\begin{equation}
C_{m}(i,j) = \sum_{k} \omega _{k}^{m} f_{k}(i,j)
\end{equation}
where $(i,j)$ is a coordinate on the feature map $f$ from the last stage of convolutional neural network, $\omega$ is the weight of last classification layer and $k$ is the $k$-th channel. 

GAP compresses a feature map from the last convolution layer into one-dimensional vector that retains the most important information. The weights of the last classification layer only activate saliency information in the feature map, which makes the seed CAM only focus on the saliency regions in the image. In addition, the structure information of the object is lost in the seed CAM. The semantic structure information can describe the correlation between feature positions, which contains clear category structure features. As shown in Figure~\ref{point_heatmap}, there is clear semantic structure information in the trained network. Therefore, it is feasible to explore semantic structure information to expand seed CAM in the inference stage. 

\begin{figure}[t]
\centering
\includegraphics[width=0.7\linewidth]{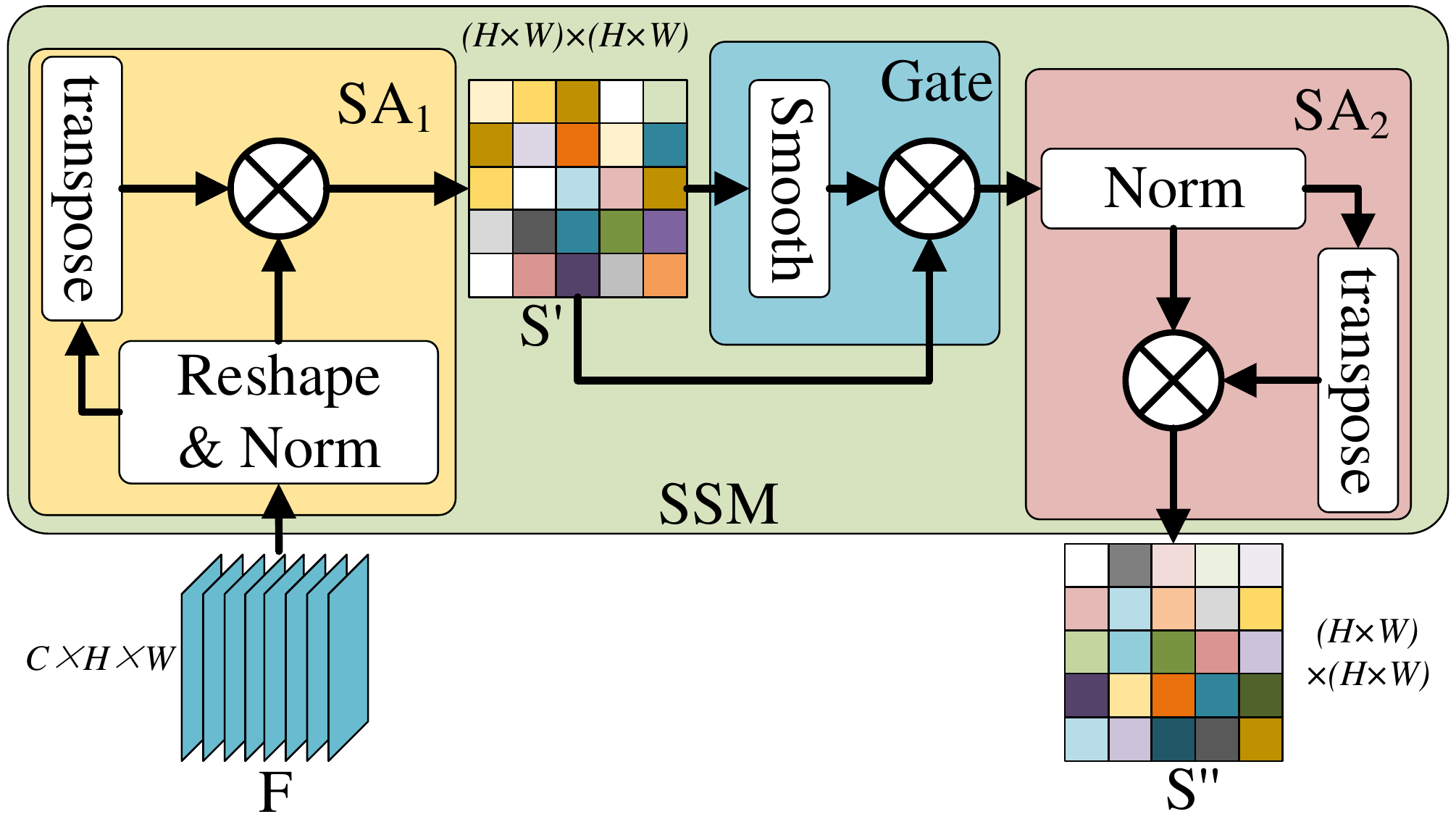} 
\caption{The details of the Semantic Structure modeling Module (SSM)}
\label{SSM}
\end{figure}

\subsection{SSM Module}
%To obtain CAMs of different scales, this paper uses the semantic structure information of different scales in stage4 and stage5 to expand seed CAM. 
Semantic structure information describes the correlation of pixels in the same category, which is very important for polish the semantic structure in CAM. This work proposes the SSM model to extract semantic structure information from the feature maps at different stages. The SSM module is composed of two self-affinity (SA) blocks and a smooth gate, and its structure is shown in the Figure~\ref{SSM}.
%The purpose of the SSM module is to extract complete semantic structure information from the feature map. Next, this section will introduce in detail how to use SSM to extract semantic structure information from feature maps at different stages. The SSM module is composed of two self-affinity (SA) blocks and a smooth gate, and its structure is shown in the Figure~\ref{SSM}.

%In order to facilitate understanding, this section uses the feature map $F_5$ from stage5 as an example to introduce the process of the SSM model.

%This paper use for reference the multi-scale fusion method to update and fuse the original CAM using the semantic structure information from different stages. Next, we will detail how to obtain the semantic structure information from stages using SSM model. The SSM module process is shown in Figure \ref{SSM}. This section takes stage5 as an example to illustrate how SSM module works. 

The feature map $f$ input to the SSM module can come from different stages of the backbone. To eliminate the adverse effects caused by the singular sample data in feature map $f$, we first reshape $f \in \mathbb{R}^{C\times H \times W}$ to $\mathbb{R}^{C\times HW}$, and then normalize it to [0,1]:
\begin{equation}
f_{(i,j)} = \frac{f(i,j)}{\sqrt[2]{\sum_{i=1}^{H}\sum_{j=1}^{W}f(i,j)}}  
\end{equation}
where $C$, $H$ and $W$ represent the channel, width and height of the feature map, respectively. Next, the first SA block captures coarse semantic structure information $S^{'}$ from the feature map $f$ by the following equation:
\begin{equation}
S^{'}=f_{a}^{T}\cdot f_{b} -\varepsilon 
\end{equation}
where $a$ and $b$ represent the index of features respectively, $\varepsilon \in \mathbb{R}^{HW\times HW}$ is an identity matrix. To reduce the impact of noise on semantic structure information, SSM introduces a smooth gate. The $tanh$ is used to obtain smooth weight of the coarse semantic structure information $S^{'}$. Next, the smooth weight is used to polish $S^{'}$:
\begin{equation}
S_{new}^{'}=tanh(S^{'} )\cdot S^{'}
\end{equation}

It is worth noting that the single SA block cannot completely cover the regions of the object. Thus, SSM leverages the SA block again to further extract more detailed semantic structure information. After the smooth gate, the semantic structure information $S_{new}^{'}$ $\in \mathbb{R}^{HW\times HW}$can be regarded as feature map, where the feature representation of each pixel is a $1\times HW$ vector. The second SA block first normalizes $S_{new}^{'}$ to [0,1]:
\begin{equation}
\tilde{S}_{(i,j)} =\frac{S_{new}^{'}(i,j)}{ {\textstyle \sum_{i=1}^{H}} {\textstyle \sum_{j=1}^{W}} S_{new}^{'}(i,j) } \label{norm}
\end{equation}
And then, the second SA block explores relations between $\tilde{S}_a$ and $\tilde{S}_b$: 
\begin{equation}
S^{''}=\phi (\tilde{S}_{a}^{T}\cdot \tilde{S}_{b} -\varepsilon) 
\end{equation}
where $\phi$ is the function ReLU. $S^{''}$ is normalized in the same way as Eq. \ref{norm}. The semantic structure information $S^{''}$ from the feature map is obtained by the SSM module. The discovered semantic structure information can be regarded as a prior knowledge to guide the process of expanding seed CAM.

\begin{algorithm}[tb]
\caption{The proposed algorithm SSA}
\label{algorithm}
\textbf{Input}: Seed CAM $C\in \mathbb{R}^{N\times H \times W}$; Feature maps $F_4\in \mathbb{R}^{C_{4}\times H \times W}$ and $F_5\in \mathbb{R}^{C_{5}\times H \times W}$ from Stage 4 and Stage 5, respectively;\\
\textbf{Output}: Final CAM $CP$
\begin{algorithmic}[1] 
\STATE Using SSM to uncover semantic structure information $S_4\in \mathbb{R}^{HW\times HW}$ in $F_4$;
\STATE Reshape $C$ to $\mathbb{R}^{N\times HW}$;
\STATE Expand $C$ with $S_4$ to get $CP_4\in \mathbb{R}^{N\times HW}$;
\STATE Reshape $CP_4$ to $\mathbb{R}^{N\times H\times W}$;
\STATE Using SSM to uncover semantic structure information $S_5\in \mathbb{R}^{HW\times HW}$ in $F_5$ ;
\STATE Expand $C$ with $S_5$ to get $CP_5\in \mathbb{R}^{N\times HW}$;
\STATE Reshape $CP_5$ to $\mathbb{R}^{N\times H\times W}$;
\STATE Fuse $CP_4$ and $CP_5$ to get the final CAM $CP$ by using Eq.\ref{fuse};
\RETURN $CP$
\end{algorithmic}
\end{algorithm}

\subsection{Semantic Structure Aware Inference}
The seed CAM only focuses on salient regions of object, which loses the structural information of the object. It is worth noting that stage 4 and stage 5 contain clear and different-scale semantic structure information, respectively. To supplement the structural information of objects in the image, SSA extracts different scales semantic structure information from different stages of the backbone to expand seed CAM. Thus, SSA extracts different scales semantic structure information from $F_4\in \mathbb{R}^{C_{4}\times H \times W}$ and $F_5\in \mathbb{R}^{C_{5}\times H \times W}$ to expand the seed CAM. First, the seed CAM $C\in \mathbb{R}^{N\times H \times W}$ is generated from the deep convolutional network, and reshaped to $\mathbb{R}^{N\times HW}$. Then SSM is introduced to extract the semantic structure information $S_4\in \mathbb{R}^{HW\times HW}$ and $S_5\in \mathbb{R}^{HW\times HW}$ from $F_4$ and $F_5$, respectively. Next, $S_4$ and $S_5$ are explored to expand seed CAM:

%Algorithm \ref{algorithm} and Figure \ref{zong} shows detailed process of SSA.

%The seed CAM can only focal the local regions of saliency and lose the structural information. As we all know, in deep convolutional networks, shallow networks pay more attention to texture and edge features in images, while deep networks pay more attention to semantic features. The visualization results in Figure \ref{point_heatmap} show that stage3 contains a lot of noise, and the semantic structure information in stage4 and stage5 is clearer.
\begin{equation}
CP_{4} = \delta (C\odot S_{4})\ \ \&  \ \ CP_{5} = \delta(C\odot S_{5})
\end{equation}
where $\odot$ represents the matrix multiplication and $\delta$ is a reshape function to reshape the results to $\mathbb{R}^{N\times H\times W}$. To further improve the accuracy of CAM, SSA fuses expansion CAM with different scales of semantic structure information to get the final CAM $CP\in \mathbb{R}^{N\times H \times W}$:
\begin{equation}
CP = CP_{4}\oplus CP_{5} 
\label{fuse}
\end{equation}
where $\oplus$ represents element-wise sum. 

With SSA, the seed CAM integrates different scales of semantic structure information, making the final CAM cover a more comprehensive regions, and clearer object structure features. It is very important for weakly supervised vision tasks.

\section{Experiments}
This section conducts experiments to evaluate the proposed SSA on two weakly supervised visual tasks: weakly supervised object localization (WSOL) and weakly supervised semantic segmentation (WSSS) with image-level class labels.\\
\textbf{Common Settings.} All experimental systems are based on Pytorch \cite{paszke2019pytorch}. Both tasks are only annotated with image-level labes for training. All models are pre-trained on ILSVRC \cite{russakovsky2015imagenet}. 

\subsection{WSOL Settings}
\textbf{Datasets.} To evaluate the SSA in WSOL task, extensive experiments are conducted on two publicly available benchmarks including CUB-200-2011 \cite{wah2011caltech} and ILSVRC \cite{russakovsky2015imagenet}. The CUB-200-2011 is a fine-grained bird dataset with 200 categories, where the training set has 5,994 images and test set has 5,794 images. The ILSVRC consists of 1,000 classes, while has 1.2 million images for the training set and 50,000 images for the validation set. In addition to image-level labels in WSOL task, the bounding box labels are also required. For ILSVRC, SEM \cite{zhang2020rethinking} provides pixel level annotation of 44,270 images for evaluating IOU curve, which is divided into the validation set (23,150 images) and the test set (21,120 images).\\
\textbf{Training Details and Metrics.} For metrics, we use two kinds of metrics to evaluate bounding box and mask from the CAM. Following the previous works \cite{zhang2018self,pan2021unveiling}, the location performance is measured by TOP1/TOP5 error and known ground-truth class (GT-known). GT-known judges the map as correct when the predicted bounding boxes have over 50\% intersection over union (IoU) with the ground-truth boxes. The mask performance is measured by IoU curve following SEM \cite{zhang2020rethinking}. The IoU curve is obtained by calculating the IoU scores between the foreground pixels and the ground-truth masks under different thresholds. High IoU scores and threshold value indicates high-quality maps. We verify the performance of SSA on VGG16 \cite{simonyan2014very} and Inception V3 \cite{szegedy2016rethinking} network respectively. For data augmentation, the input images are resized to 256$\times$ 256 pixels and randomly cropped to 224$\times$ 224 pixels. For classification, we average the scores from the softmax layer with 10 crops. 

\begin{table}[t]
		% increase table row spacing, adjust to taste
		\renewcommand\arraystretch{1.1}
		\caption{Ablation study based on the number of SA block in SSM module. \textbf{N-SA} denotes the number of the SA block.}
	    \label{numself_table}
		\centering
		\begin{tabular}{l|c|c|c|c}
			\hline
			\multirow{2}*{Method} & \multirow{2}*{N-SA} & \multicolumn{3}{c}{Error rate (\%)}\\
			\cline{3-5}
			&&TOP-1 & TOP-5 & GT-known\\
			\hline
			CAM \cite{zhou2016learning}& - & 55.9 & 47.8 & 44.0\\
			CAM + SSA & $1$ & 45.1 & 33.2 & 29.2\\
			CAM + SSA & $2$ & \textbf{44.9} & \textbf{33.0} & \textbf{29.1}\\
			CAM + SSA & $3$ & 49.2 & 38.2 & 34.2\\
			\hline
			SPA + SCG \cite{zhou2016learning}& - & 40.49 & 28.4 & 23.0\\
			SPA + SSA & $1$ & 38.6 & 25.1 & 19.8\\
			SPA + SSA & $2$ & \textbf{38.1} & \textbf{24.9} & \textbf{19.4}\\
			SPA + SSA & $3$ & 41.8 & 29.8 & 24.6 \\
			\hline
	\end{tabular}
\end{table}

\subsection{WSSS Settings}
\textbf{Datasts.} To verify the validity of SSA in WSSS, experiments are conducted on the PASCAL VOC 2012 \cite{everingham2010pascal} dataset, which contains 1,464 images in the training set, 1,449 images in the validation set and 1,456 images in the test set. Following the common practice, the augmented dataset \cite{hariharan2011semantic} expands the training set from the original data to 10,582 images, which have image-level labels. This dataset contains pixel-level labels for fully supervised segmentation, but we use the image-level class labels for training and pixel-level labels for evaluation.\\
\textbf{Training Details and Metrics.} To evaluate the performance of weakly semantic segmentation, the widely used mean intersection-over-union (mIoU) is introduced. Following the previous work \cite{zhang2020causal}, ResNet50 \cite{he2016deep} is used to generate the initial seed, and then IRNet \cite{ahn2019weakly} is used to generate the pseudo labels. For fair competition, DeepLab-v2-ResNet101 \cite{chen2017deeplab} is used in the final segmentation stage. In the CAM generation phase, to eliminate the noise caused by flip inference, we first use the $HardTanh$ function with range $[0,1]$ to obtain the significant map $SM_4$ and $SM_5$ in $CP_4$ and $CP_5$, then use $SM_4$ to guide the generation of $CP5$, and $SM5$ to guide the generation of $CP_4$. The initial learning rate is set to 0.1 with 10 epoch for training classification network, while 5 epoch for training IRNet. In the final segmentation, the batch size is set to 10, and the learning rate is set to 2.5e-4 with 30K.

\begin{figure}[t]
\centering
\includegraphics[width=0.7\linewidth]{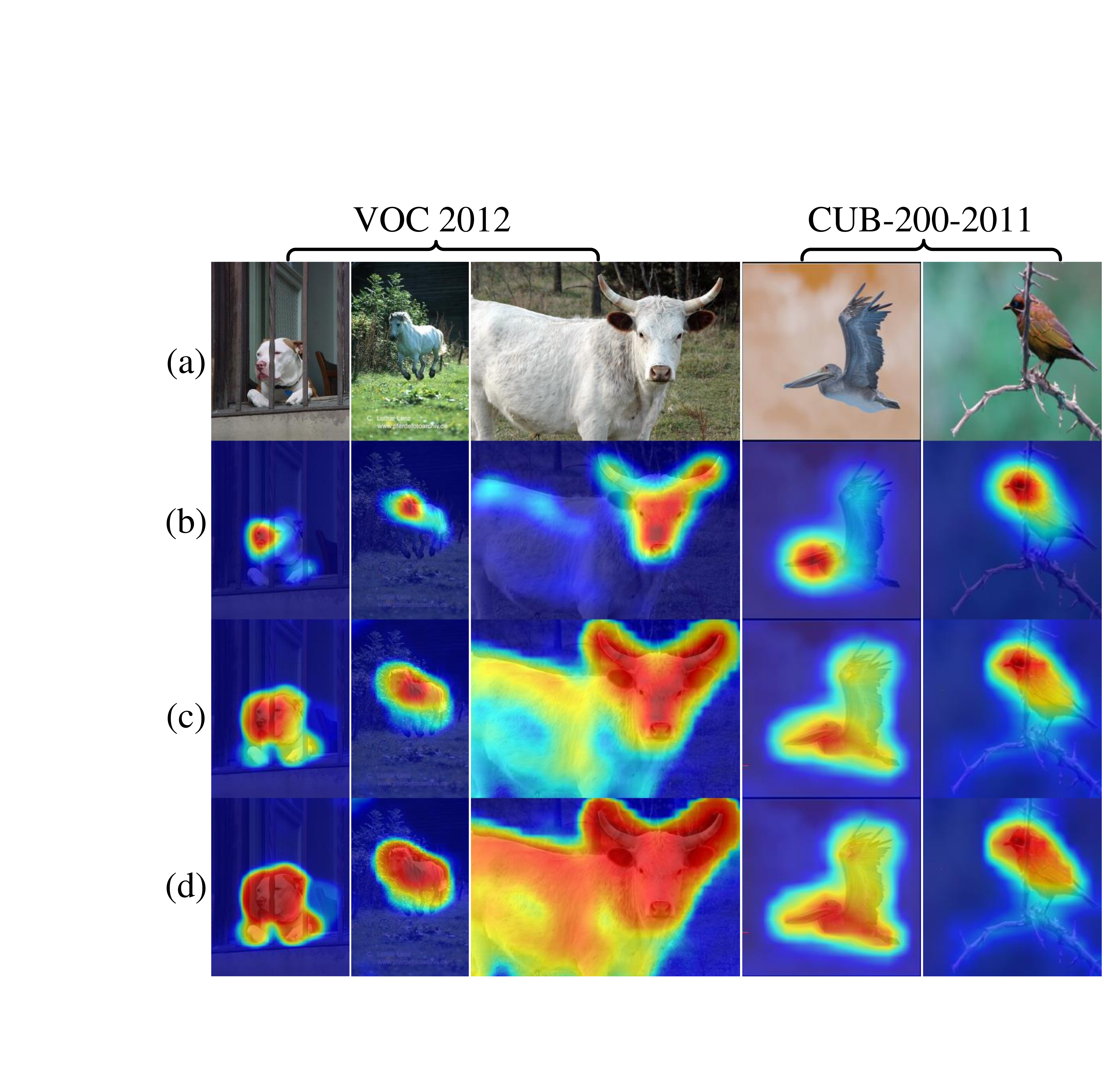} 
\caption{Visualization results of SSA with different numbers of SA-blocks. (a) input image, (b) seed CAM, (c) and (d) are the results of SSA after using SA block once and twice, respectively.}
\label{num_self}
\end{figure}

\subsection{Ablation Study}
This section conducts a series of ablation study on the CUB-200-2011, ILSVRC and PASCAL VOC 2012 datasets. The results are obtained by using VGG16 as the backbone on CUB-200-2011 and using ResNet50 as the backbone on PASCAL VOC 2012. VGG16 and Inception V3 are used as the backbone on ILSVRC respectively. We first verify the effectiveness of the proposed SSM module, then verify the strength of the semantic structure information at different stages, and finally verify the universality of SSA. 

%First, we test the influence of the SSM module on the extraction of semantic structure information, secondly test the strength of semantic structure information in different stages, and finally test the universality of CAM Smooth.

\textbf{For SSM.} In the SSM module, the SA block is proposed to capture the semantic structure in the features. Thus, the ablation study is conducted to show the superiority of the SSM module. First, experiments are conducted to verify the influence of the number of SA blocks (N-SA) in the SSM model. The number of SA is tuned within $[1, 2, 3]$. Experimental results based on CAM \cite{zhou2016learning} and SPA \cite{pan2021unveiling} are shown in Table \ref{numself_table}. It can be easily seen that when the SSM module uses two SA blocks, the best performance is achieved, which indicates that the SSM module obtains more detailed semantic structure information from the feature map. To visually observe the influence of N-SA on the final CAM, we performed a visualization experiment. The result are shown in Figure \ref{num_self}. The visualization results show that when N-SA is $2$, the regions of the final CAM is more accurate than when the number of SA blocks is $1$. For example, in the third column images of Fig.\ref{num_self}, when the SA block is used only once, the CAM can only cover the head of the cow. When the SA block is used twice, CAM can cover the whole body of the cow. Experimental results show that SSM can obtain more accurate semantic structure information from the feature map after using two SA blocks.

%It can be easily seen that CAM Polish achieves optimal performance when the number of SA blocks is equal to $2$. It shows that the semantic structure information obtained is the most complete when N-SA is equal to $2$. 

\begin{figure}[t]
\centering
\includegraphics[width=0.7\linewidth]{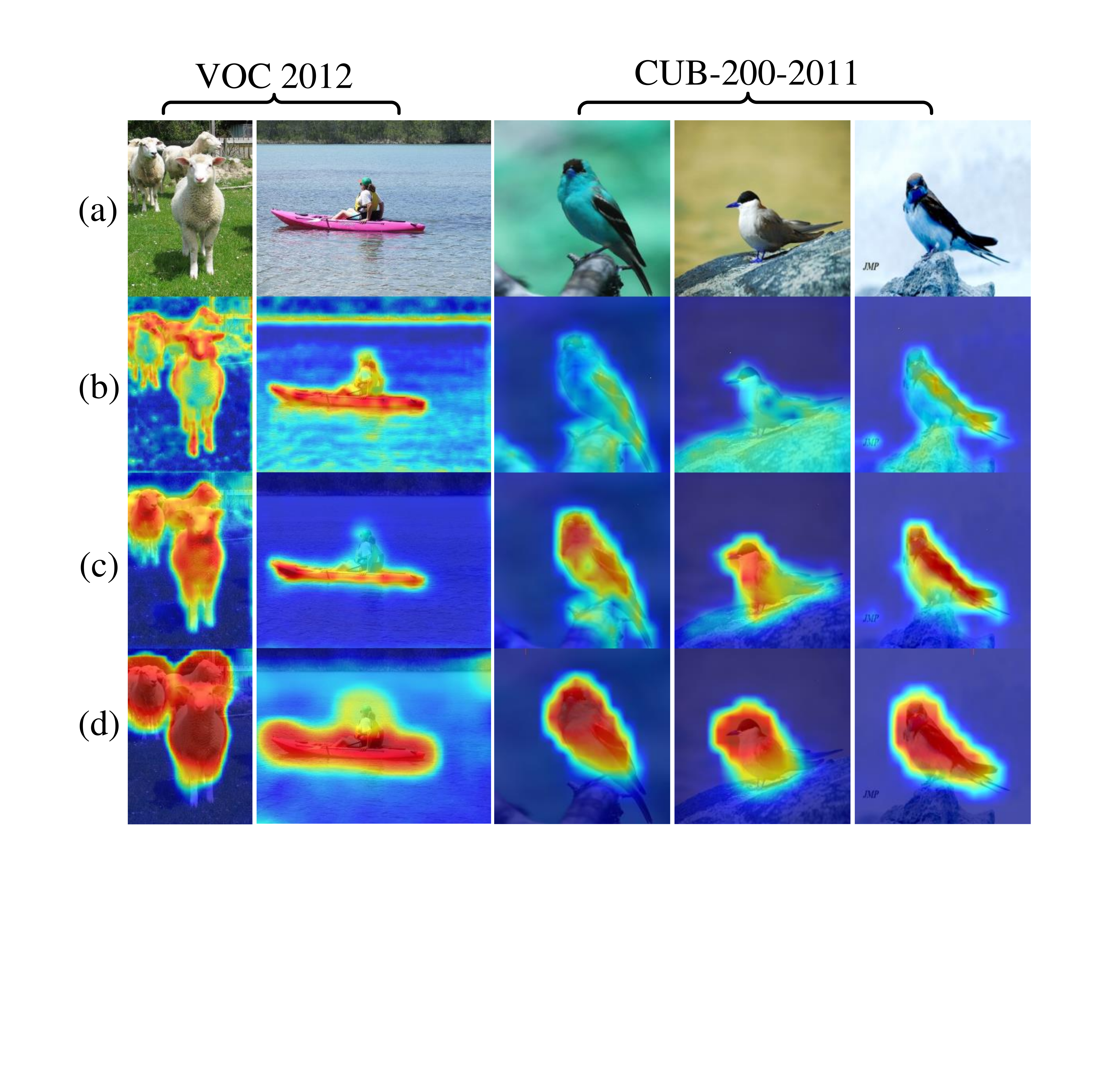} 
\caption{Visualization results of SSA with semantic structure information of different stages. (a) input image, (b) using Stage 3, (c) using Stage 4 and (d) using Stage 5}
\label{stage_cam}
\end{figure}

\textbf{For SSA.} The most important part in SSA is to explore the semantic structure information from different stages to expand the seed CAM. Therefore, which stage has more detailed semantic structure information is the focus of the next study. To verify the validity of the semantic structure information in different stages, SSA first uses the SSM module to extract the semantic structure information in different stages, then expands the seed CAM through it, and finally observes its performance in the WSOL task. SPA without SCG \cite{pan2021unveiling} serves as the benchmark for this experiment. The ablation study results are presented in Table \ref{stage_table}. It can be easily observed that the semantic structure information in stage 5 improves the model performance most. To visually observe the changes of the final CAM that integrates the semantic structure information of different stages, we conducted a visualization experiment, and the results are shown in Figure \ref{stage_cam}. The results show that stage 3 introduces a lot of noise information for final CAM, and stage 4 introduces noise for final CAM, but it can clearly cover all areas of the object. Although the final CAM obtained using stage 5 covers the largest area, it has the phenomenon of over-recognition. In order to further improve the quality of CAM, a simple method is to merge the CAM after fusing the semantic structure information of different stages. The results are show in Table \ref{stage_table}. It can be seen that the fusion of stage 4 and stage 5 can further improve the performance of SSA. Due to the excessive noise in stage 3, the combination of stage 3 cannot improve the quality of CAM.

\begin{table}[t]
		% increase table row spacing, adjust to taste
	
		\renewcommand\arraystretch{1.1}
		\caption{Ablation study of SSA on CUB-200-2011. \textbf{Stage} indicates the semantic structure information of which stage is used to expand the seed CAM.}
	    \label{stage_table}
		\centering
		\begin{threeparttable}
		\begin{tabular}{c|c|c|c}
			\hline
			\multirow{2}*{Stage} & \multicolumn{3}{c}{Error rate (\%)}\\
			\cline{2-4}
			&TOP-1 & TOP-5 & GT-known\\
			\hline
			$\dagger \dagger$ & 49.0 & 38.5  & 34.1 \\
			Stage 3 & 66.4 & 59.2 & 56.5 \\
			Stage 4 & 44.8 & 33.0 & 28.1\\
			Stage 5 & \textbf{40.1} & \textbf{27.3} & \textbf{21.9}\\
			\hline
			Stage 3 + Stage 4 & 49.7 & 39.3 & 34.7\\
			Stage 3 + Stage 5 & 40.0 & 27.0 & 21.6\\
			Stage 4 + Stage 5 & \textbf{38.1} & \textbf{24.9} & \textbf{19.4}\\
			Stage 3 + Stage 4 + Stage5 & 39.0 & 26.1 & 20.6 \\
			\hline
	\end{tabular}
	\begin{tablenotes}\tiny
        \footnotesize
        \item $\dagger \dagger$ denotes the benchmark SPA without SCG
      \end{tablenotes}
    \end{threeparttable}
    
\end{table}

\begin{figure}[t]
\centering
\includegraphics[width=0.7\linewidth]{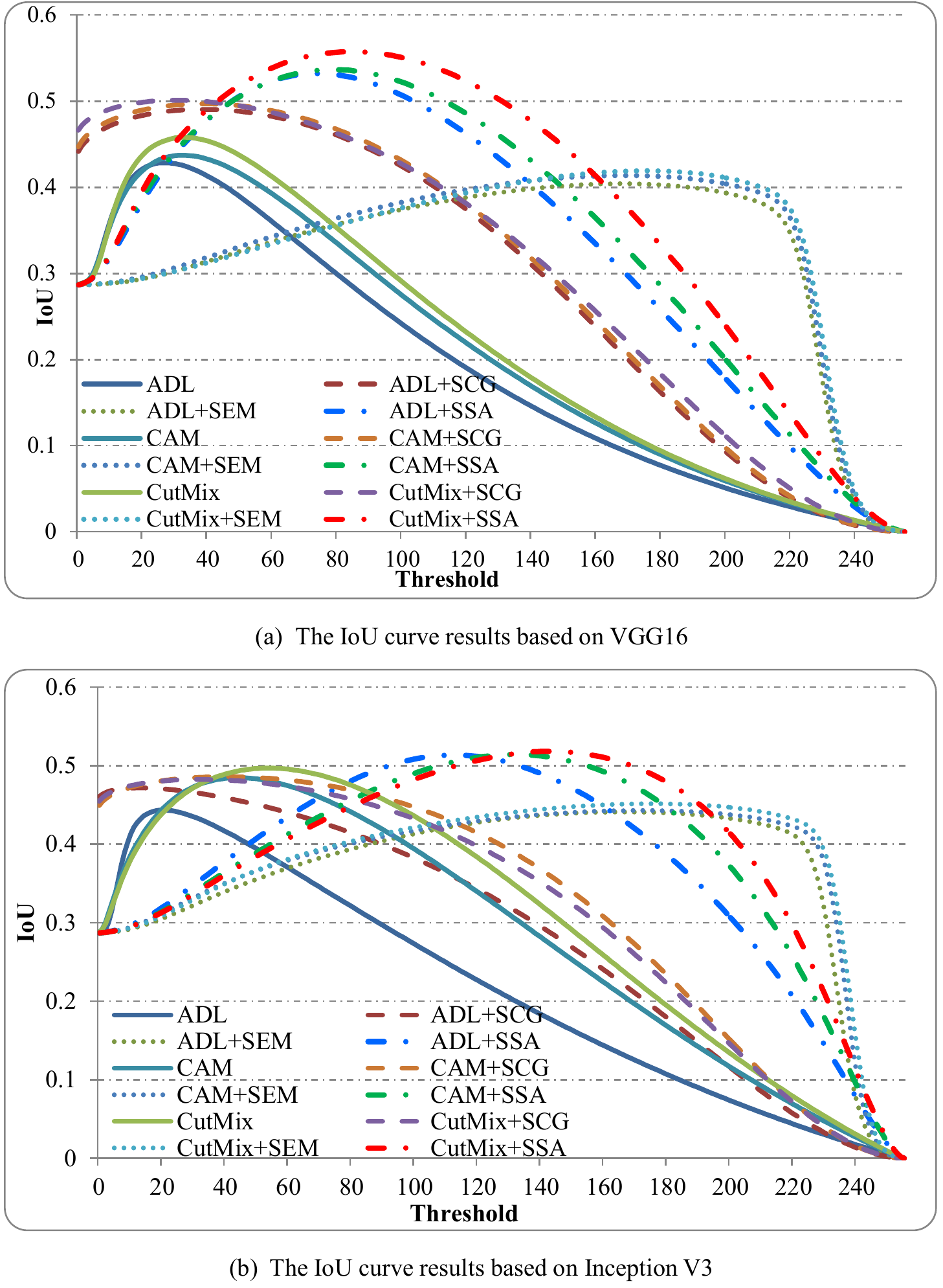} 
\caption{Compared results in terms of IoU curve on the ILSVRC dataset.}
\label{curve}
\vspace{-1.5em}
\end{figure}

\textbf{For the universality and superiority of SSA.} To verify the universality of SSA, IoU curve is used to evaluate the performance of SSA on the ILSVRC dataset. The compared methods include ADL \cite{choe2019attention}, CAM \cite{zhou2016learning} and CutMix \cite{yun2019cutmix} based on VGG16 and Inception V3. In addition, the superiority of SSA is verified by comparing SSA, SEM \cite{zhang2020rethinking} and SCG \cite{pan2021unveiling}, which expand CAM in the inference step. The IoU curve results are shown in Figure \ref{curve}. It can be seen that SSA greatly improves the quality of CAM, and the change of backbones does not affect the performance of SSA. Experimental results show that although the accuracy of CAM obtained by SCG is improved, the mask generated by CAM has the worst visual effect. The mask obtained by SEM has a good visual effect, but its localization accuracy is very poor. Compared with SCG and SEM, the mask obtained by SSA has not only higher localization accuracy, but also better visual effects. High-quality CAM generated mask is characterized by high clarity and high localization accuracy, which are important for weakly supervised vision tasks. Therefore, in order to further improve the quality of CAM, SSA is the best choice in the inference step.

\subsection{State-of-the-Art Comparisons}
\textbf{Comparison with SOTA in WSOL.} To verify the effectiveness of the proposed SSA in object localization, experiments are carried out on the CUB-200-2011 and ILSVRC datasets. SSA based on SPA \cite{pan2021unveiling} and CAM \cite{zhou2016learning} are compared with SPG \cite{zhang2018self}, DANet \cite{xue2019danet}, SEM \cite{zhang2020rethinking}, I$^2$C \cite{zhang2020inter}, etc. Table \ref{CUB-SOTA} shows the experimental results of SSA and SOTA methods on the CUB-200-2011 test set. By using Ineptionv3 as the backbone, SSA surpasses all the compared methods and achieves TOP1 localization error rate of 44.9\% in the original CAM. With VGG16, SSA achieves the TOP1 localization error rate of 44.9\% in the original CAM, and TOP1 localization error rate of 38.1\% based on the stronger classification network (SPA), which surpasses all the baselines. Table \ref{ILSVRC} reports the results on the ILSVRC test set. It can be seen that SSA achieves TOP1 localization error rate of 50.7\%, supassing all the baselines. Based on the same network, compared with SCG, SSA reduces the TOP1 localization error rate by 2.4\% and 0.4\% on the CUB-200-2011 and ILSVRC data sets, respectively. It shows that SSA can be seamlessly integrated into any CNN-based network to improve the quality of CAM, thereby improving the performance of object localization without cost.

\begin{table}[t]
		% increase table row spacing, adjust to taste
	
		\renewcommand\arraystretch{1.1}
		\caption{Comparison with state-of-the-arts on the CUB-200-2011 test set.}
	    \label{CUB-SOTA}
		\centering
		\begin{threeparttable}
		\begin{tabular}{l|c|c|c|c}
			\hline
			\multirow{2}*{Method} & \multirow{2}*{Backbone} & \multicolumn{3}{c}{Error rate (\%)}\\
			\cline{3-5}
			&&TOP-1 & TOP-5 & GT-known\\
			\hline
			CAM \cite{zhou2016learning} & GoogLeNet & 58.9 & 49.3 & 44.9\\
			SPG \cite{zhang2018self} & GoogLeNet & 53.4 & 42.8 & -\\
			CAM \cite{zhou2016learning} $\dagger$ & InceptionV3 & 53.8 & 42.8 & 38.3\\
			DANet \cite{xue2019danet} & InceptionV3 & 50.6 & 39.5 & 33.0\\
			SEM \cite{zhang2020rethinking}& GoogLeNet & 47.0 & - & 30.0\\
			ADL \cite{choe2019attention}& InceptionV3 & 47.0 & - & -\\
			SPA+SCG \cite{pan2021unveiling} & InceptionV3 & 46.4 & 33.5 & 27.9\\
			\hline
			CAM + SSA  & InceptionV3 & \textbf{44.9} & \textbf{31.5} & \textbf{25.5}\\
			\hline
			CAM \cite{zhou2016learning}& VGG16 & 55.9 & 47.8 & 44.0\\
			ACoL \cite{zhang2018adversarial}& VGG16 & 54.1 & 43.5 & 45.9\\
			SPG \cite{zhang2018self} & VGG16 & 51.1 & 42.2 & 41.1\\
			ADL \cite{choe2019attention}& VGG16 & 47.6 & - & -\\
			DANet \cite{xue2019danet} & VGG16 & 47.5 & 38.0 & 32.3\\
			I$^2$C \cite{zhang2020inter} & VGG16 & 44.0 & 31.6 & -\\
			MEIL \cite{mai2020erasing} & VGG16 & 42.5 & - & -\\
			SPA+SCG \cite{pan2021unveiling} $\dagger$ & VGG16 & 40.5 & 28.4 & 22.9\\
			\hline
			CAM + SSA  & VGG16 & 44.9  & 33.0  & 29.1 \\
			SPA + SSA  & VGG16 & \textbf{38.1}  & \textbf{24.9}  & \textbf{19.4}  \\
			\hline
	\end{tabular}
	\begin{tablenotes}\tiny
        \footnotesize
        \item $\dagger$ denotes the results implemented by us
      \end{tablenotes}
    \end{threeparttable}
\end{table}

\begin{table}[t]
		% increase table row spacing, adjust to taste
		\renewcommand\arraystretch{1.1}
		\caption{Comparison with state-of-the-arts on the ILSVRC test set.}
	    \label{ILSVRC}
		\centering

		\begin{tabular}{l|c|c|c|c}
			\hline
			\multirow{2}*{Method} & \multirow{2}*{Backbone} & \multicolumn{3}{c}{Error rate (\%)}\\
			\cline{3-5}
			&&TOP-1 & TOP-5 & GT-known\\
			\hline
			%Backprop \cite{simonyan2014deep} & VGG16 & 61.1 & 51.45 & -\\
			CAM \cite{zhou2016learning}& VGG16 & 57.2 & 45.1 & -\\
			CutMix \cite{yun2019cutmix}& VGG16 & 56.6 & - & -\\
			SEM \cite{zhang2020rethinking}& VGG16 & 55.4 & - & 39.2\\
			ADL \cite{choe2019attention}& VGG16 & 55.1 & - & -\\
			ACoL \cite{zhang2018adversarial}& VGG16 & 54.2 & 40.6 & 37.0\\
			MEIL \cite{mai2020erasing} & VGG16 & 53.2 & - & -\\
			I$^2$C \cite{zhang2020inter} & VGG16 & 52.6 & 41.5 & 36.1\\
			SPA+SCG \cite{pan2021unveiling} $\dagger$ & VGG16 & 51.1 & 39.5 & 35.1\\
			\hline
			SPA + SSA & VGG16 & \textbf{50.7}  & \textbf{39.0}  & \textbf{34.4}  \\
			\hline
	\end{tabular}
\end{table}

\textbf{Comparison with SOTA in WSSS.} To verify the effectiveness of SSA in multi-label tasks, we conducted experiments on PASCAL VOC 2012 dataset for the WSSS task. The SSA is applied in the CAM generation step. Following the previous work \cite{zhang2020causal}, IRNet is used for post-processing to generate pseudo-labels. Our method is compared with CIAN \cite{fan2020cian}, OAA \cite{jiang2019integral}, SEAM \cite{wang2020self}, CONTA \cite{zhang2020causal} etc. Experiments results are presented in Table \ref{WSSS}. It can be observed that SSA achieves the best result 67.4\% and 67.9\% mIoU on the val and test sets respectively, which surpasses all the baselines. Specifically, the performance of SSA surpasses methods such as BoxSup \cite{dai2015boxsup} and SDI \cite{khoreva2017simple} that use bounding boxes as supervisory information. In addition, compared with IRNet, SSA improves the performance on the val and test data sets by 3.9\% and 3.1\% mIoU, respectively. It shows that the proposed SSA can help CAM to cover more correct areas, and obtain high-quality pseudo-labels.

\section{Conclusion}
This paper proposes a new semantic structure aware inference (SSA) for weakly pixel-wise dense predictions, which explores different scales of semantic structure information to generate high-quality CAM in inference. Specifically, The Semantic Structure modeling Module (SSM) is developed to explore the semantic structure on different backbone stages. Then, the semantic structure is leveraged to expand the seed CAM, and finally the expand CAMs on different stages are merged to obtain the final CAM. The proposed SSA can be applied to any CNN-based method to improve the quality of CAM. SSA achieves superior performance on WSOL and WSSS tasks, respectively. However, the generalization ability of semantic structure information is insufficient. Thus how to improve the generalization ability of the semantic structure information is the focus of our future research.

%In this study, we use the semantic structure information in different stages to update the original CAM and apply it to the pixel-wise Dense predictions task. First, we discover the semantic structure information in the feature map through visualization experiments. Then use the SSM module to extract the semantic structure information in the feature map, and finally use it to update the original CAM. Since the receptive field of each stage in the deep convolutional network is different, the semantic structure information of different stages is different. To obtain high-quality CAM, we use the semantic structure information in multiple stages to update the original CAM, and finally merge them. Finally, verify the effectiveness of SSA on WSOL and WSSS tasks. The experimental results show that our method obtains the best results in both tasks, which proves the effectiveness and superiority of the method.

\begin{table}[t]
		% increase table row spacing, adjust to taste
		\renewcommand\arraystretch{1.0}
		\centering
		\caption{Comparison with state-of-the-arts on the PASCAL VOC 2012 val and test sets.}
	    \label{WSSS}
		\begin{threeparttable}
		\begin{tabular}{lccc}
			\hline
			Method & Sup. & val & test\\
			\hline
			\hline
			\emph{Fully supervised} \\
			FCN \cite{long2015fully}& $\mathcal{P.}$  & -& 62.2\\
			Deeplab \cite{chen2017deeplab} & $\mathcal{P.}$  & 67.7 & 70.3\\
			\hline
			\emph{Weakly supervised}\\
			BoxSup \cite{dai2015boxsup} & $\mathcal{B.}$  & 62.0 & 64.6\\
			SDI \cite{khoreva2017simple} & $\mathcal{B.}$  & 65.7 & 67.5\\
			MCIS \cite{sun2020mining}& $\mathcal{I.\space S.\space W.}$  & 67.7 & 67.5\\
			OAA++$^+$ \cite{jiang2021online}& $\mathcal{I.\space S.}$  & 66.1 & 67.2\\
			AttnBN \cite{li2019attention}& $\mathcal{I.\space S.}$  & 62.1 & 63.0\\
			OAA \cite{jiang2019integral}& $\mathcal{I.\space S.}$  & 63.9 & 65.6\\
			FickeNet \cite{lee2019ficklenet}& $\mathcal{I.\space S.}$  & 64.9 & 65.3\\
			MCIS \cite{sun2020mining}& $\mathcal{I.\space S.}$  & 66.2 & 66.9\\
			CIAN \cite{fan2020cian}& $\mathcal{I.\space S.}$  & 64.3 & 65.3\\
			SEC \cite{kolesnikov2016seed} & $\mathcal{I.}$  & 50.7 & 51.7\\
			AE-PSL \cite{wei2017object}& $\mathcal{I.}$  & 55.0 & 55.7\\
			IRNet \cite{ahn2019weakly}& $\mathcal{I.}$  & 63.5 & 64.8\\
			SSDD \cite{shimoda2019self} & $\mathcal{I.}$  & 64.9 & 65.5\\
			SEAM \cite{wang2020self} & $\mathcal{I.}$  & 64.5 & 65.7\\
			SC-CAM \cite{chang2020weakly}& $\mathcal{I.}$  & 66.1 & 65.9\\
			CONTA \cite{zhang2020causal}& $\mathcal{I.}$  & 66.1 & 66.7\\
			\hline
			IRNet + SSA & $\mathcal{I.}$  & \textbf{67.4} & \textbf{67.9}\\
			\hline
	\end{tabular}
	\begin{tablenotes}
        \footnotesize
        \item $\mathcal{P.}$-Pixel level mask, $\mathcal{B.}$-Box, $\mathcal{I.}$-Image level class, $\mathcal{W.}$-Web
      \end{tablenotes}
    \end{threeparttable}
    
\end{table}

%Bibliography
\bibliographystyle{unsrt}  
\bibliography{references}

\end{document}